\documentclass[twocolumn]{svjour3}          
\smartqed  
\RequirePackage{fix-cm}
\usepackage{graphicx}
\usepackage{amsmath}
\usepackage{threeparttable}
\usepackage{caption} 
\usepackage{amsmath}
\usepackage{tabularx}
\usepackage{url}
\usepackage{floatrow}
\usepackage{amssymb}
\usepackage{bbding}
\usepackage[switch,pagewise]{lineno} 

\title{PPI-NET: End-to-End Parametric Primitive Inference}
\author{Liang Wang \and Xiaogang Wang\textsuperscript{\Envelope}  }
\institute{Liang Wang \at College of Computer and Information Science, Southwest University,China.
\and Xiaogang Wang\textsuperscript{\Envelope} corresponding author\at College of Computer and Information Science, Southwest University,China.}
\date{ }

\begin{document}
\maketitle
\begin{abstract}
In engineering applications, line, circle, arc, and point are collectively referred to as primitives, and they play a crucial role in path planning, simulation analysis, and manufacturing. When designing CAD models, engineers typically start by sketching the model's orthographic view on paper or a whiteboard and then translate the design intent into a CAD program. Although this design method is powerful, it often involves challenging and repetitive tasks, requiring engineers to perform numerous similar operations in each design. To address this conversion process, we propose an efficient and accurate end-to-end method that avoids the inefficiency and error accumulation issues associated with using auto-regressive models to infer parametric primitives from hand-drawn sketch images. Since our model samples match the representation format of standard CAD software, they can be imported into CAD software for solving, editing, and applied to downstream design tasks. 
\end{abstract}

\section{Introduction}
\label{sec:1}
Parametric computer-aided design (CAD) is a method of designing, analyzing, and optimizing using computer-aided design tools, which is widely used in fields such as mechanical engineering and aerospace. CAD tools can help professionals design faster and more accurately, thereby improving the efficiency and quality of product development and providing a structured representation for downstream tasks such as simulation and manufacturing.

\begin{figure*}[t]
    \centering
    \includegraphics[width=\textwidth,trim= 0 0 0 0,clip]{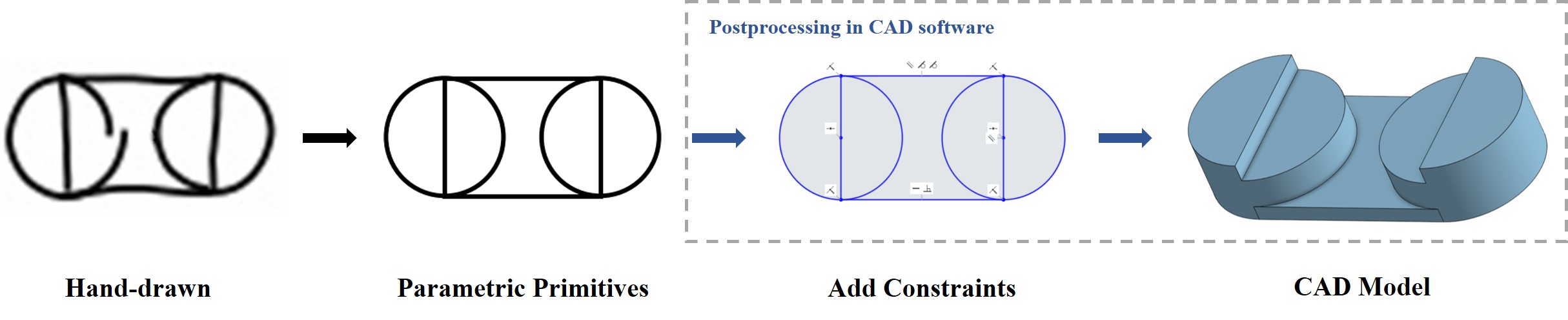}
    \caption{CAD model generation is conditioned on a hand-drawn sketch image. Hand-drawn sketch images as input, we feed them into our the image primitive network to generate parametric primitives. Import the primitives into CAD software, by adding constraints, adjust the positions of primitives to create a coherent sketch, final extrusion the sketch to generate a 3D model.}
    \label{fig:2Dhand_draw_2_3Dmodel}
\end{figure*}

In practical operation, parametric CAD begins with a two-dimensional geometric specification, which is typically referred to as a "sketch" in the CAD community \cite{shah2001designing,wu2021deepcad}. The sketch comprises geometric primitives (line, circle, arc, point). By importing the primitives into CAD software, we can edit the sketch by adding constraints and adjusting the positions of the primitives. With further CAD operations on the sketch, such as extrusion, a 3D model can be generated. This process involves gradually adjusting the constraints and parameters between the primitives to accurately express the design intent and generate the desired 3D model. \figurename~\ref{fig:2Dhand_draw_2_3Dmodel}, illustrates a functionality achieved through our method.  While this design methodology is robust, it is typically a challenging and repetitive process, with an engineer performing dozens of similar operations in each design. Accurately learning to predict this pattern will reduce the number of repetitive manual procedures and increase work efficiency. Additionally, engineers typically start visualizing structures by hand-drawn sketches and implementing primitives on professional software. The automatic and reliable conversion of hand-drawn sketches or similar noisy inputs (such as 3D scans) into parametric and editable models remains a highly desirable feature.

Object detection is a fundamental task in computer vision. Classic convolutional-based object detection algorithms have made significant progress \cite{bochkovskiy2020yolov4,lin2017focal,ren2015faster}. Although such algorithms often include hand-designed components such as anchor generation and non-maximum suppression (NMS), they have produced state-of-the-art detection models such as DyHead \cite{dai2021dynamic}, SwinV2 \cite{liu2022swin}, and InternImage \cite{wang2022internimage}, which have been demonstrated in the COCO test development leaderboard.

Compared to traditional detection algorithms, DETR \cite{DETR} is a new transformer-based detection algorithm. It eliminates the need for hand-designed components and achieves performance comparable to optimized classic detectors such as Faster RCNN \cite{ren2015faster}. Unlike previous detectors, DETR \cite{DETR} models object detection as a set prediction task and assigns labels through bipartite graph matching. It uses learnable queries to detect the presence of objects and combines features from image feature maps.

Despite its good performance, DETR \cite{DETR} suffers from slow convergence during training and unclear interpretation of the queries. To address such issues, many approaches have been proposed, such as introducing deformable attention \cite{zhu2020deformable}, decoupling position and content information \cite{meng2021conditional}, Anchor DETR \cite{wang2022anchor} directly treat 2D reference points as queries to perform cross-attention, and so on. Recently, DAB-DETR \cite{liu2022dab} proposes to formulate the DETR \cite{DETR} queries as dynamic anchor boxes (DAB), which bridges the gap between classical anchor-based detectors and DETR-like detectors. DN-DETR \cite{li2022dn} and DINO \cite{zhang2022dino} further solve the instability issue of bipartite matching by introducing the denoising (DN) technique.


\begin{table}[h]
\centering
\begin{tabular}{c|c|c|c|c|c|c}
\hline
\textbf{Line}   & $x_1$ & $y_1$ & $x_2$ & $y_2$ &  $0$ &  $0$ \\ \hline
\textbf{Circle} & $x$ & $y$ & $r$ & 0  & 0  &  0 \\ \hline
\textbf{Arc}    & $x_1$ & $y_1$ & $x_{mid}$ & $y_{mid}$ & $x_2$ & $y_2$ \\ \hline
\textbf{Point}  & $x$ & $y$ & 0  & 0  & 0  & 0  \\ \hline
\end{tabular}
\caption{We convert the native Onshape numerical parameters to the below forms for modeling. Subscripts $1$, $mid$, and $2$ indicate the start, mid, and end point, respectively. Value 0 means padding.} 
\label{tab:table_one} 
\end{table}

All previous detection methods detect the object category associated with the bounding box, as well as the four parameters $(x,y,w,h)$ of the box, with fixed parameters that do not involve changes in parameter lengths. In our task, as shown in Table \ref{tab:table_one}, line has four parameters, circle has three parameters, arc has six parameters, and point has two parameters. Meanwhile, unused parameters for each primitive are simply set to be $0$. The parameter length of the primitive varies dynamically according to its type, which is unsuitable for the anchor-based DETR method described above.

In this work, we introduce the PPI-NET network, a detection network that, after training, detects geometric primitives to synthesize coherent CAD sketches.  The image primitive model uses the original Transformer and a denoising module in DN-DETR \cite{li2022dn} to infer the parameters of primitives end-to-end. In denoising module, we flipped ground truth labels and added noisy parameters of primitives in the Transformer decoder layers to help stabilize bipartite matching during training for accelerating the convergence of DETR \cite{DETR}. It is trained on the filtered version of the SketchGraphs dataset \cite{seff2020sketchgraphs} by Vitruvion \cite{seff2021vitruvion}, which contains information on four primitive types, with a total of 1.7 million samples. We make the following contributions:
\begin{enumerate}
  \item We convert real hand-drawn sketches into editable forms within CAD software, enabling designers to communicate and collaborate more easily with team members, clients, and suppliers.

  \item We can bind different parameters to various types of primitives by employing primitives' dynamic parametric inference. This provides us with an effective means to optimize the representation of primitives and cater to real world usage scenarios.
    \item We propose an end-to-end network called PPI-Net for inferring parametric primitives from hand-drawn sketch images, which provides users with a more flexible and manipulable data representation.
    \item Our method has achieved better results compared with Vitruvion \cite{seff2021vitruvion} in both qualitative and quantitative aspects. 
\end{enumerate}
\section{Related work}
\label{sec:2}
\subsection{Parametric primitive inference}
Parameterized primitive fitting is a long-standing problem in the field of geometric processing. Researchers have extensively explored methods for detecting or fitting parameterized feature curves, such as Bézier and B-spline, often using least squares representation \cite{flory2008constrained,li2011design,wang2006fitting}. ParSeNet \cite{sharma2020parsenet} decomposes a 3D point cloud into a set of parametric surface patches. PIE-Net\cite{wang2020pie} infers three-dimensional parameterized curves by analyzing edge point clouds. Point2cly \cite{uy2022point2cyl} segments point clouds into patches belonging to different primitives and performs primitive type classification and parameter regression for each patch. On the other hand, ComplexGen \cite{guo2022complexgen} adopts a boundary representation (B-Rep) approach to simultaneously recover corner points, curves, and surfaces along with their topological constraints. These methods share the common characteristic of taking three-dimensional point cloud data as input.
In contrast, Vitruvion \cite{seff2021vitruvion} extracts parameterized primitives from real hand-drawn sketches, while Sketchgen \cite{para2021sketchgen} utilizes parameterized primitives for constructing two-dimensional sketches, considering them as the first step towards three-dimensional entities.
\begin{figure*}[t]
    \centering
    \includegraphics[width=0.93\textwidth,trim= 0 0 0 0,clip]{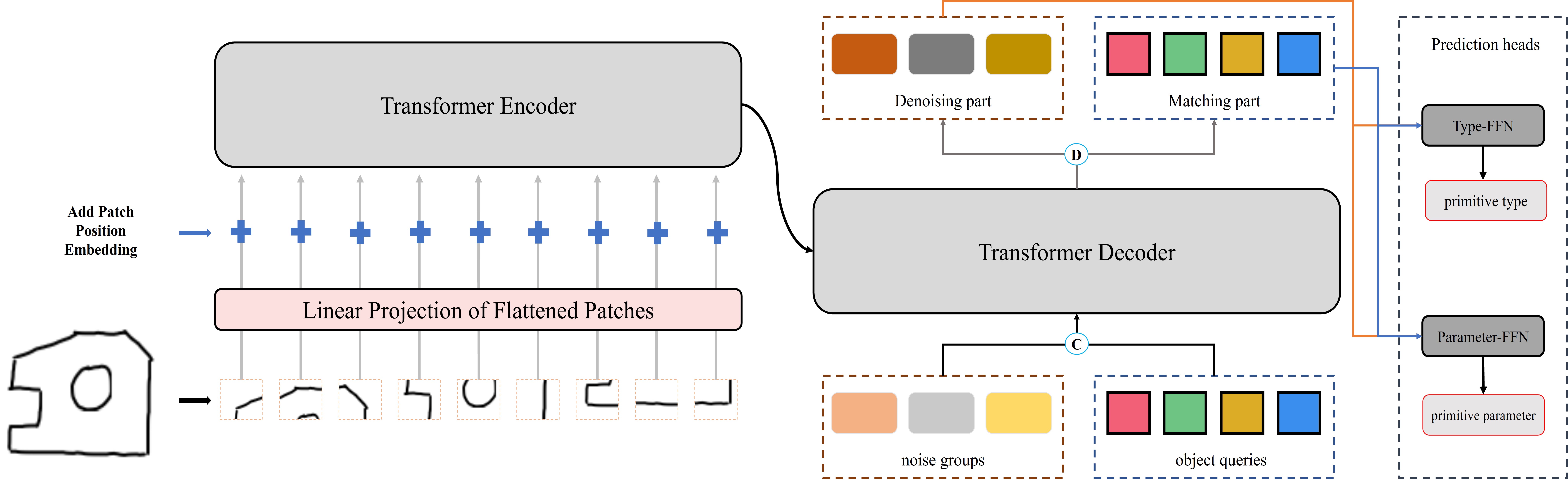}
    \caption{\textbf{Image Primitive Model Pipeline.} Input a hand-drawn sketch image, and the network predicts the type and parameters of the primitive. "C" means cat the noise groups and object queries. On the contrary, "D" means decoupling the noise groups and object queries. } 
    \label{fig:Image Primitive Model Pipeline}
\end{figure*}
\subsection{CAD sketch generation}
The SketchGraphs dataset \cite{seff2020sketchgraphs} was recently introduced to target the underlying relational structure and construction operations found in mechanical design software. Specifically, the dataset provides ground truth geometric constraint graphs, thus the initial design steps, for millions of real world CAD sketches created in Onshape. 
KDD Willis et al.\cite{willis2021engineering} employing sketch generation models, it becomes possible to transform 2D sketch primitives into 3D entities and ultimately compose a final 3D model by combining multiple entities. \cite{ganin2021computer,para2021sketchgen,seff2021vitruvion} all utilize auto-regressive models to model the primitives and the constraint relationships between them. However, this approach cannot directly output results end-to-end and is easily influenced by the previous time steps' outputs. Additionally, neither \cite{ganin2021computer} nor Sketchgen \cite{para2021sketchgen} addresses the problem of recovering primitives from hand-drawn sketches. In contrast, Vitruvion \cite{seff2021vitruvion} takes hand-drawn sketches as input to the network and expands the application scenarios. Similar to our method, this input approach allows for better reconstructing primitives from hand-drawn sketches and application in a broader range of domains.

\subsection{Vector graphics generation}

Vector graphics are used extensively in commercial software to enable the resolution independent design of fonts, logos, animations, and illustrations. In recent years, generative models have been successfully applied to the field of vector graphics. For instance, Sketch-RNN \cite{ha2017neural} learns to generate vector graphics of generic categories (e.g., dogs, buses) by training on sequences of human drawing strokes.  
BézierGAN \cite{chen2020airfoil} is a method that utilizes a generative adversarial network (GAN) \cite{goodfellow2020generative} to synthesize smooth curves. Im2Vec \cite{reddy2021im2vec}, on the other hand, employs differentiable rendering techniques \cite{lopes2019learned} for vector graphics generation. DeepSVG \cite{carlier2020deepsvg} focuses on modeling SVG icons and demonstrates flexible editing capabilities through latent space interpolation. Additionally, tracing programs have been widely used for decades to convert raster drawings to vector graphics \cite{selinger2003potrace}.
The process of Vectorization \cite{egiazarian2020deep} involves dividing the cleaned image into multiple patches. For each patch, the initial parameters of the primitives, which can be either line segments or quadratic Bézier curves, are evaluated. Finally, Vectorization \cite{egiazarian2020deep} combines and refines the primitives of each patch to obtain the vector graphic. Our method, similar to the image-conditional version of Vitruvion \cite{seff2021vitruvion}, takes hand-drawn raster images as input and generates vector graphics.
%
\subsection{Transformer-based models}
 Technically, our work is related to the Transformer network \cite{vaswani2017attention}, which was introduced as an attention-based building block for many natural language processing tasks \cite{devlin2018bert}. The Transformer model utilizes a self-attention mechanism to capture global information in both the encoder and decoder, enabling parallel computation between different positions in the input sequence. Due to its remarkable scalability and performance, the Transformer network has been widely applied in image processing tasks \cite{DETR,li2022dn,zhang2022dino} and other types of data processing \cite{carlier2020deepsvg}. Some relevant works parallel ours, such as DETR \cite{DETR} and DN-DETR \cite{li2022dn}, also rely on the Transformer network for object detection in images. DN-DETR \cite{li2022dn}, in particular, optimizes DETR \cite{DETR} by introducing denoising groups for faster convergence and higher accuracy. Free2CAD \cite{Li:2022:Free2CAD} utilizes a Transformer model to build a sequence-to-sequence model, which takes a given sequence of strokes as input and groups the strokes corresponding to individual CAD operations. These groups are combined with geometric fitting of the operation parameters to ultimately generate a CAD model. Additionally, Vitruvion \cite{seff2021vitruvion} is related to our work as well. It is a Transformer-based network that extracts primitives from hand-drawn sketches, aiming to alleviate the time and effort required for engineers to convert them into CAD programs.
\section{Method}
\label{sec:3}
The pipeline of our proposed PPI-Net network is illustrated in \figurename~\ref{fig:Image Primitive Model Pipeline}. Given an input image, a detection module detects the four types of primitives implicitly embedded in the image and extracts their parameterized information as object queries.
\subsection{Overview}
Similar to DETR \cite{DETR}, our architecture contains Transformer encoders and Transformer decoders. On the encoder side, the image features are extracted with MLP and fed into the Transformer encoder with positional encodings to attain refined image features. On the decoder side, object queries are fed into the decoder to search for objects through cross-attention. 

We denote decoder queries as $\textbf{q}=\{{q_0,q_1,...,q_{N-1}\}}$ and the output of the Transformer decoder as $o=\{o_0,o_1,...,o_{N-1}\}$. We also use $F$ and $A$ to denote the refined image features after the Transformer encoder and the attention mask derived based on the denoising task design. We can formulate our method as follows.
\begin{equation}
\label{eq1}
 o=D(\textbf{q},F\ |\ A) ,
\end{equation}where $D$ denotes the Transformer decoder.

There are two parts of decoder queries. One is the matching part. The inputs of this part are learnable primitive parameters, which are treated in the same way as in DETR. The matching part adopts bipartite graph matching and learns to approximate the ground truth primitive parameter-label pairs with matched decoder outputs. The other is the denoising part. The inputs of this part are noised ground-truth (GT) primitive parameter-label pairs, which are called $GT objects$ in the rest of the paper. The outputs of the denoising part aim to reconstruct GT objects.

In the following, we abuse the notations to denote the
denoising part as $\textbf{q}=\{{q_0,q_1,...,q_{N-1}\}}$ and the matching part as $\textbf{Q}=\{Q_0,Q_1,...,Q_{N-1}\}$. So the formulation of our method becomes
\begin{equation}
\label{eq2}
o=D(\textbf{q},\textbf{Q},F\ |\ A),
\end{equation}
In order to enhance the effectiveness of denoising, we suggest incorporating multiple variations of the noised GT objects during the denoising process. Additionally, we employ an attention mask to safeguard against information leakage from the denoising phase to the matching phase, as well as between different versions of the same GT object with noise.

\subsection{Denoising}
We collect all GT objects for each image and add random noises to their parameter and class labels. To maximize the utility of denoising learning, we use multiple noised versions for each GT object. 
For label noise, we use label flipping, which means we randomly flip some GT labels to other labels. The purpose of label flipping is to encourage the model to predict the GT labels based on the noised primitives' parameter, thus improving its ability to capture the relationship between labels and primitives. We have a hyper-parameter $\gamma$  to control the ratio of labels to flip. 

For primitives' parameter noise, which is obtained by adding noise of standard normal distribution to the corresponding GT primitive parameters of each GT primitive type, $\lambda$ to control the ratio of noise to add, as 
$q_{parameter}=(GT_{parameter}+\lambda *noise)* mask_{GT_{type}}$. 

The losses of denoising part are $l1$ loss and Chamfer Distance ($CD$) loss for primitive of parameters and focal loss \cite{lin2017focal} for class labels as in DAB-DETR \cite{liu2022dab}. In our method, we employ a function $\delta$(·) to represent the noised GT objects. Thus, each query in the denoising part can be expressed as $q_{m}=\delta(GT_{m})$, where $GT_{m}$ corresponds to the $m$-th GT object.
\subsection{Attention Mask}

The inclusion of an attention mask is vital in our model. It has been observed from the results presented in Table \ref{tab:table_Five}, that the absence of an attention mask during denoising training hinders performance rather than enhancing it. To introduce attention mask, we need to first divide the noised GT objects into groups. Each group is a noised version of all GT objects.The denoising part becomes
\begin{equation}
\label{eq1}
q=\{{g_0,g_1,...,g_{p-1}}\},
\end{equation}
where $g_p$ is defined as the $p$-th noise group. Each noisie groups contains $N$ queries where $N$ is set to be significantly larger than the typical number of primitives in an image. So we have
\begin{equation}
\label{eq1}
g_p=\{{q_0^p,q_1^p,...,q_{N-1}^p}\},
\end{equation} where $q_p^n=\delta(GT_n)$.
The purpose of the attention mask is to prevent information leakage. There are two types of potential information leakage. One is that the matching part may see the noised
GT objects and easily predict GT objects. The other is that one noised version of a GT object may see another version. Therefore, our attention mask is to make sure the matching part cannot see the denoising part and the denoising groups
cannot see each other.

We use $A = {[a_{ij}]}_{W\times W}$ to denote the attention mask where $W = P\times N$, $P$ is the numbers of
groups. We let the first $P\times N$ rows and columns to represent the denoising part and the latter to represent the matching part. $a_{ij}$ = 1 means the $i$-th query cannot see the $j$-th query and $a_{ij}$ = 0 otherwise. 
Note whether the denoising part can see the matching
part or not will not influence the performance. This is because the queries in the matching part are learned queries that do not contain any information about the GT primitive. 
\begin{equation}
    a_{ij} = \begin{cases}
                0, & \text{if } P*N \leq(i,j)\leq (P+1)*N, \\
                1, & otherwise.
           \end{cases}
\end{equation}
\subsection{Embedding}
The decoder embedding is specified as label embedding in our model to support both primitive parameter denoising and label denoising. We also append an indicator to label embedding. The indicator is 1 if a query belongs to the denoising
part and 0 otherwise. We embed the noise label as a content query. We denote $P_i = (p_1,p_2,p_3,p_4,p_5,p_6)$ as the noise parameter information of the $i$-th primitive , $p_1,p_2,p_3,p_4,p_5,p_6 \in [0,1]$. $D$ is the dimension of decoder embeddings and positional queries. Given a primitive parameter $P_i$, its positional query $V_i$ is generated by:
\begin{equation}
\label{eq1}
V_i = MLP(PE(P_i)),
\end{equation}
where $PE$ means positional encoding to generate sinusoidal embeddings from float numbers and the parameters of MLP are shared across all layers. As $P_i$ has six elements , we overload the $PE$ operator here:
\begin{equation}
\label{eq2}
\begin{split}
   PE(P_i) &= CAT(PE(p_1,p_2,p_3,p_4,p_5,p_6)),
\end{split}
\end{equation}
The notion $CAT$ means concatenation function. In our implementations, the positional encoding function $PE$ maps a float to a vector with ${D/2}$ dimensions as: 
$R \rightarrow R^{D/2}$. Hence the function MLP projects a ${3D}$ dimensional vector into $D$ dimensions: MLP: $R^{3D} \rightarrow R^{D}$. The MLP module has two submodules, each composed of a linear layer and a ReLU activation, and the feature reduction is conducted at the first linear layer.
\subsection{Image primitive model}
The image primitive model is inspired by DETR \cite{DETR} and applies the denoising part of DN-DETR \cite{li2022dn} to accelerate the convergence of DETR \cite{DETR}. We hope it can accurately recover parameterized primitives from hand-drawn sketches. 
In real CAD design, engineers usually draw orthogonal projection contours of the 3D model from various angles before investing time in building the CAD model. With this starting point, accurately inferring parameterized primitives from images is a highly sought-after feature in the CAD field but also can significantly reduce the workload for designers to convert rough sketches on paper into programs in CAD software.

\textbf{Architecture.} The model is based on an image encoder using a visual converter \cite{dosovitskiy2020image}, whose task is to generate a sequence of image representations for the decoder to perform cross-attention. We use 128 × 128 size image as input for the model. Extract size 16 from input image 16 × 16 non-overlapping square blocks are flattened to produce a sequence of 64 flattened blocks. Then, before entering the standard converter encoder, each one undergoes a linear transformation of the embedding dimension of the model (256 in this case), and adds corresponding position encoding to each image sequence. For the decoder, which is the same as DETR, we use learnable position encoding as object queries and use it as input for each attention layer. Object queries will be converted into outputs in each Decoder layer. Ultimately, they will all independently pass the Type-FFN and Parameter-FFN, predicting the corresponding primitive types and corresponding primitive parameters separately.

\textbf{Denoising \& Matching Part.} The matching part uses the same method as the original DETR \cite{DETR} and obtains the corresponding relationship between the predicted object and ground truth through bipartite matching, thus performing loss calculations. Denoising part: since we know the primitive type and primitive parameters that each query should correspond to when building the noise groups, we can directly calculate the loss without bipartite matching. Please note that the denoising part is only considered during training, removing the denoising part during the inference process and leaving only the matching part.

\textbf{Cost Matrix.} Like DETR \cite{DETR}, each image contains different primitive types and parameters, so we use bipartite matching to determine the corresponding relationship between prediction and ground truth. The cost matrix for classification is the same as DAB-DETR \cite{liu2022dab}. Treating a parameter with a length of six as the parameter to be used, and directly comparing the parameter to be used with the ground truth parameter for L1 loss as ${cost_{p}}$. The parameters to be used are sampled according to four types: line, circle, arc, and point. The CD between each type of sampling point and the sampling point of ground truth is calculated, and the type with the smallest distance is taken for the cost matrix as ${cost_{CD}}$. 
\begin{equation}
Cost\ Matrix = \omega_{c}*cost_{c} +\omega_{p}*cost_{p}+\omega_{cd}*cost_{CD},
\end{equation}
This optimal assignment is computed efficiently with the Hungarian algorithm.

\textbf{Loss.} With the bipartite matching result, the one-to-one correspondence between prediction and ground truth, we can calculate the classified loss as focal loss for primitives type, as $Loss_{c}$. Calculate the loss of primitive parameters, 
\begin{equation}
\label{eq4}
Loss_{p} = \frac{1}{K}{\sum_{i=1}^K {||\hat{y}-y||} *mask_{T_y}},
\end{equation}
$K$ is the actual number of primitives in the image, ${T_y}$ is the corresponding GT primitive type, $\hat{y}$ is the predicted primitive parameter, and $y$ is the ground truth primitive parameter. Because different types of parameters use different positions, so use the binary mask to calculate the actual loss. 

Using the $\hat{y}$ and $y$, we perform point sampling based on ${T_y}$, and calculate the CD between the sampled points, which bind different parameters to various types of primitives,
\begin{equation}
\label{eq6}
Loss_{CD} = \frac{1}{K}{\sum_{i=1}^K{CD(\hat{y},y,T_y)}},
\end{equation}
$Loss_{CD}$ represents the sum of CD for all types of primitives. 
The final loss is the weighted sum of the above loss, 
\begin{equation}
\label{eq5}
Loss = \omega_{c}*Loss_{c}+\omega_{p}*Loss_{p}+\omega_{cd}*Loss_{CD}.
\end{equation}

\section{Experiments}
\label{sec:4}

\begin{figure}[t]
    \centering
    \includegraphics[width=3.2in,height=6.8in,keepaspectratio]{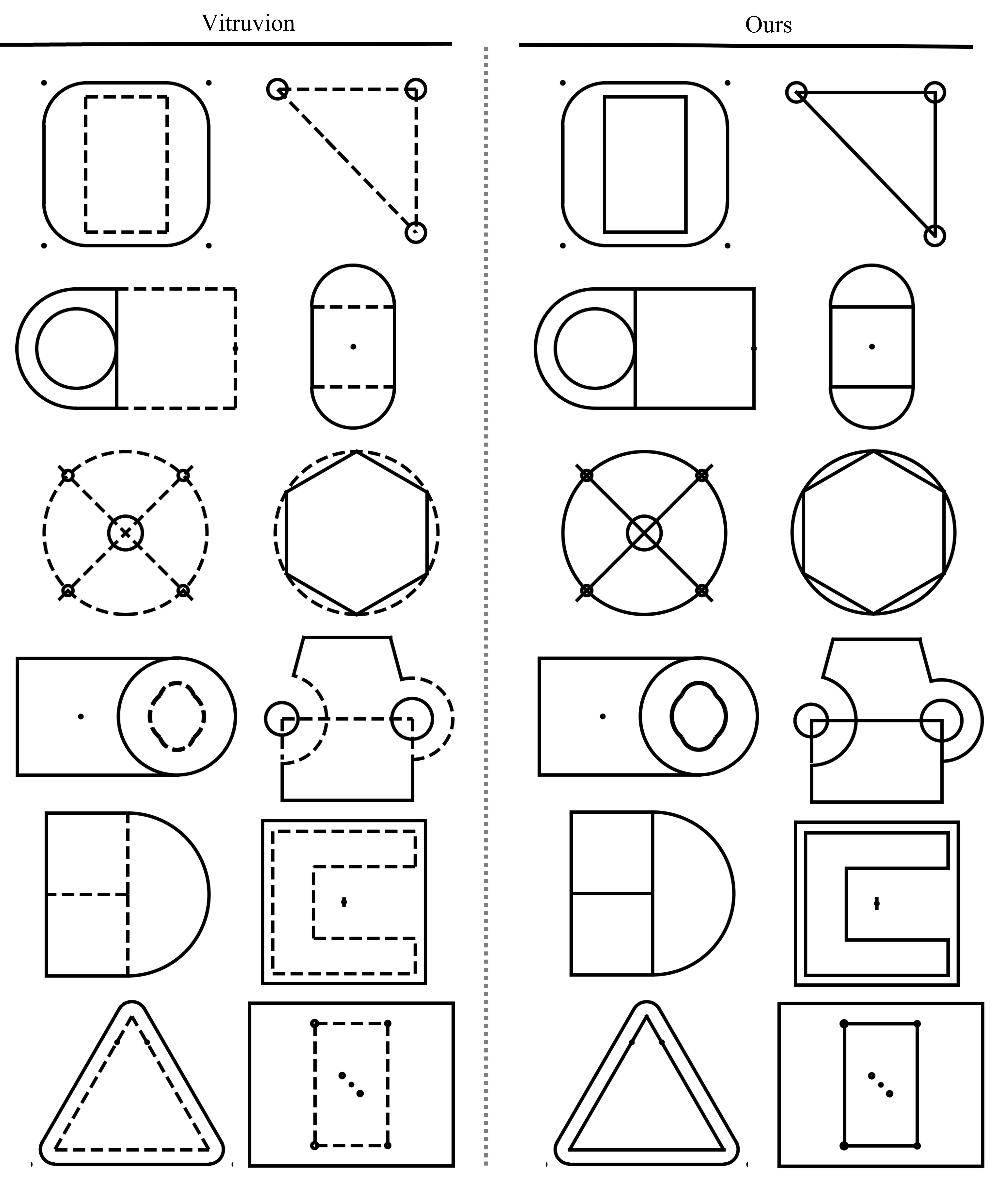}
    \caption{Each primitive has an attribute isConstruction, and when isConstruction = $True$, Vitruvion renders the primitive with a dashed line $(left)$. In contrast, our method disregards attribute isConstruction and directly renders all primitives with solid lines $(right)$.} 
    \label{fig:isConstruction}
\end{figure}
\begin{figure}[h!t]
    \centering
    \includegraphics[width=3.5in,height=6.8in,keepaspectratio]{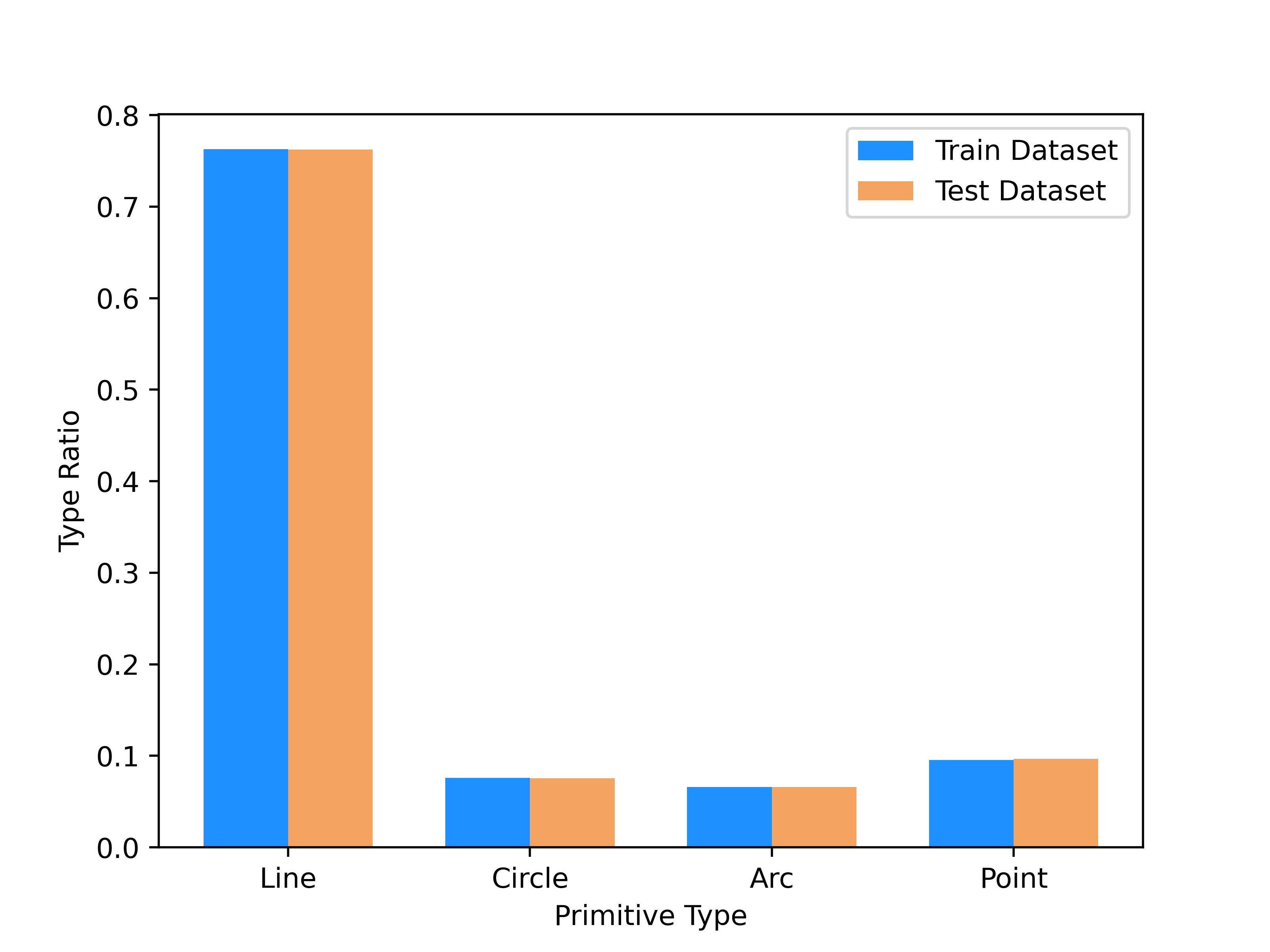}
    \caption{The proportion of each primitives type in the dataset.} 
    \label{fig:type ratio}
\end{figure}

\subsection{Training dataset}
Our model is trained on a filtered version of the SketchGraphs dataset \cite{seff2020sketchgraphs} by Vitruvion \cite{seff2021vitruvion}, which is limited to sketches composed of the four most common primitives (arc, circle, line, and point) normalized each sketch via centering and uniform rescaling, with a maximum of 16 primitives per sketch and filtered out any sketches with less than six primitives to eliminate most trivial sketches (such as simple rectangles), resulting in a collection of 1.7 million unique sketches. As shown in the \figurename~\ref{fig:type ratio}, we demonstrate the relative proportion of each primitives type in the dataset. However, unlike Vitruvion \cite{seff2021vitruvion}, we do not consider the $isConstruction$ boolean of the parametric primitive in Table \ref{tab:table_one}, which Construction or virtual geometry is employed in CAD software to aid in applying constraints to regular geometry. $IsConstruction$ specifies whether the given primitive is to be physically realized (when $False$) or simply serve as a reference for constraints (when $True$), clarifying differences through rendering as shown in \figurename~\ref{fig:isConstruction}. We randomly divided the filtered sketch set into 92.5\% training, 2.5\% validation, and 5\% testing partitions. 

\subsection{Hand-drawn noise model}
The noise model uses a zero-mean Gaussian process with a Matérn-3/2 kernel to render lines, which are rotated into the correct orientation. Arcs are rendered as paths in polar coordinates using a Matérn Gaussian process, with the angle as input and the radius modulated by a random function. No additional observation noise is introduced, apart from the standard "jitter" term. The length scale and amplitude are selected and the Gaussian process is truncated to ensure consistent scales regardless of the line or arc's length scale.
\subsection{Image process}
Before training the image primitive model, we generate samples using a hand-drawn noise model in a parallelized process on a computing cluster. Approximately 24 hours are required to generate five samples for each sketch in the dataset. During each training epoch, the model selects one random image from the five generated samples. To further augment the data during training, random affine transformations are applied to the images. These transformations include translation (up to 8 pixels), rotation (up to 10 degrees), shear (up to 10 degrees), and scaling (up to 20\%).

\subsection{Evaluation metrics}
\label{metrics}
We implemented four metrics to measure the performance of the image primitive model: $(1)$ type accuracy, which measures the primitive type classification performance; $(2)$ The Chamfer Distance (CD), which expected predicted parameters are matched with the ground truth parameters of the corresponding type to minimize the distance between them; $(3)$ precision, $(4)$ recall,
both of them are originated from the field of object detection, where the detected objects are matched with the ground truth objects to measure the matching results. In object detection, the performance of an algorithm is evaluated by comparing the detected objects with the ground truth objects.

\begin{figure*}[h!t]
    \centering
    \includegraphics[width=\textwidth,height=\textheight,keepaspectratio]{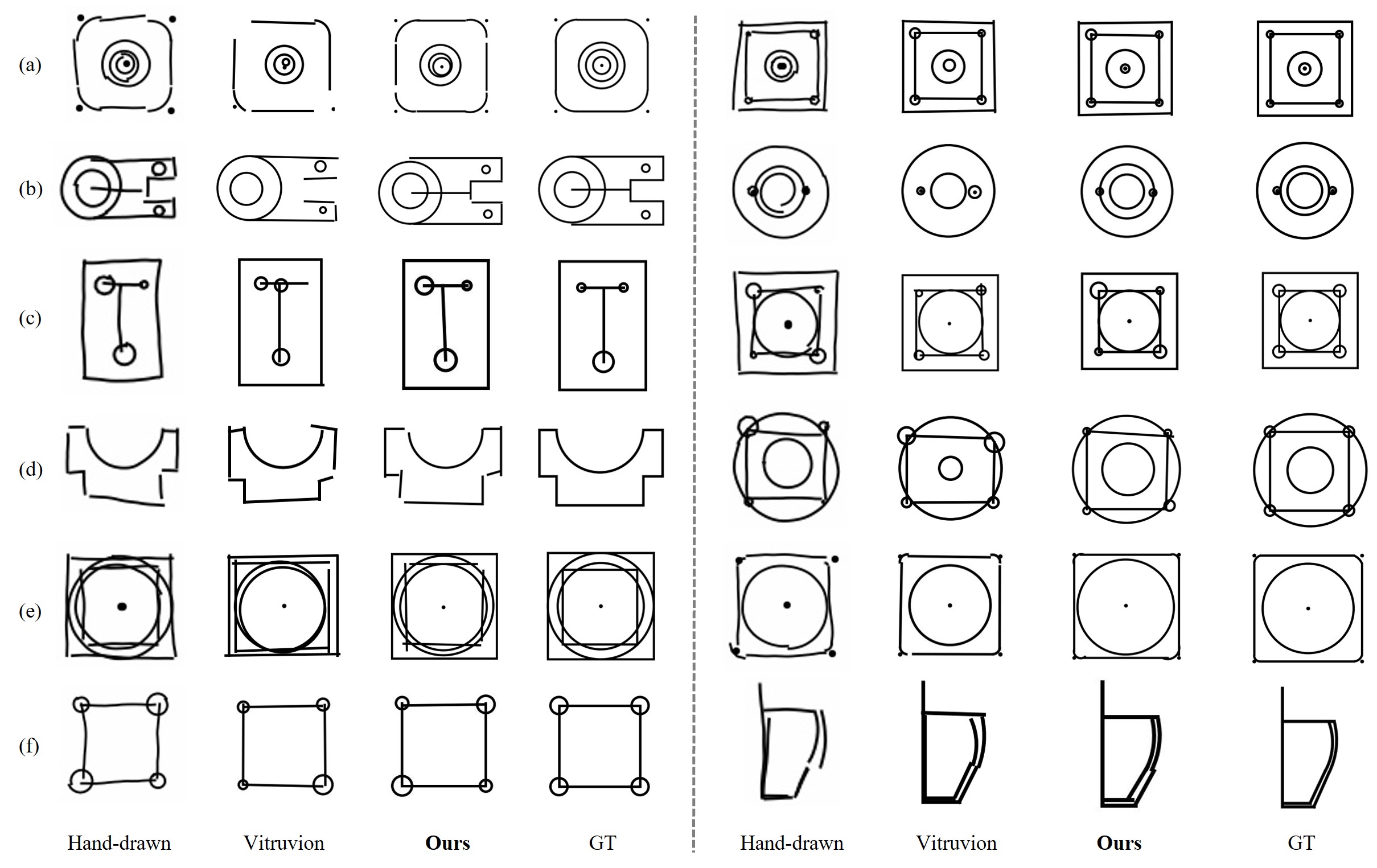}
    \caption{Compare with Vitruvion \cite{seff2021vitruvion}. Raster images of real hand-drawn sketches are input to the image primitive model.} 
    \label{fig:contrast}
\end{figure*}
The type accuracy is calculated as: 
\begin{equation}
{Type\_acc} = \frac{1}{K}{\sum_{i=1}^{K}1(t_i={\hat{t}_i})},
\end{equation}
where $K$ is the actual number of primitives contained in the image, $\hat{t_i}$ is the predicted primitive type, and ${t_i}$ is the ground truth primitive type. 
\begin{equation}
{Precision} = \frac{TP}{TP+FP},
\end{equation}
\begin{equation}
{Recall} = \frac{TP}{TP+FN},
\end{equation}
We use the probabilities of the primitive types as the confidence scores and the CD as one of the evaluation metrics. 
\begin{equation}
TP= {score_{confidence} > \tau_{con}}\  \&\  {CD < {\tau_{CD}}},
\end{equation}
\begin{equation}
FP= {score_{confidence} > \tau_{con}}\  \&\  {CD > {\tau_{CD}}},
\end{equation}
\begin{equation}
FN= {score_{confidence} < \tau_{con}}\  \&\  {CD > {\tau_{CD}}},
\end{equation}
In this case, we use $\tau_{con} = 0.50$ and $\tau_{CD} = 0.40$ as the threshold values for the evaluation metrics.
\subsection{Parameter setting}
The training was conducted on a server with four Nvidia 3090 GPUs, and our model was trained for 250 epochs, which took 72 hours. There is still room for improvement as the iterations continue. Our model is configured with standard Transformer encoder and decoder components, with six layers each. We set $\gamma = 0.4$ to control the label flipping and $\lambda = 0.3$ to control the noise parameter in denoising part. We have also incorporated three sets of denoising groups and the weight parameters for loss are set to $\omega_{c} = 2$, $\omega_{p} = 2$, and $\omega_{cd} = 5 $. All models were trained with $N = 20$ decoder query slots. The models were trained using the Adam optimizer with “decoupled weight decay regularization $"$ \cite{loshchilov2017decoupled}, with the learning rate set according to a “one-cycle $"$ learning rate schedule \cite{smith2019super}. The initial learning rate was set to 3e-5 (at reference batch size 128, scaled linearly with the total batch size). The batch size was set 256 / GPU for the image primitive model.

\subsection{Comparisons to the state-of-the-art}
Currently, there is relatively limited research on the task of inference parametric primitives from real hand-drawn images. To the best of our knowledge, in 2021, the Vitruvion \cite{seff2021vitruvion} introduced this task for the $first$ time and employed an auto-regressive model to address it. Given the significance of this task, we aim to qualitatively and quantitatively compare our research with Vitruvion \cite{seff2021vitruvion}.

Under the same input conditions, \figurename~\ref{fig:contrast} showcases our model's ability to detect primitives, in Table \ref{tab:table_Four} has been evaluated through quantitative comparisons with the image to primitive model of Vitruvion \cite{seff2021vitruvion}. Our model demonstrates outstanding performance in these aspects. Due to its auto-regressive nature, Vitruvion \cite{seff2021vitruvion} may result in missing primitives when the end token appears in the output sequence, as seen in (\figurename~\ref{fig:contrast}($a,b$)). However, our method focuses more on inferring the implied primitive sizes in the input image, as shown in (\figurename~\ref{fig:contrast}($c,d$)). Our method and Vitruvion exhibit gaps between primitives since each primitive is treated as an independent entity without considering their relational constraints. However, our method performs better in this aspect, as seen in (\figurename~\ref{fig:contrast}($e,f$)). 

\begin{figure}[t]
    \centering
    \includegraphics[width=3.5in,height=6.8in,keepaspectratio]{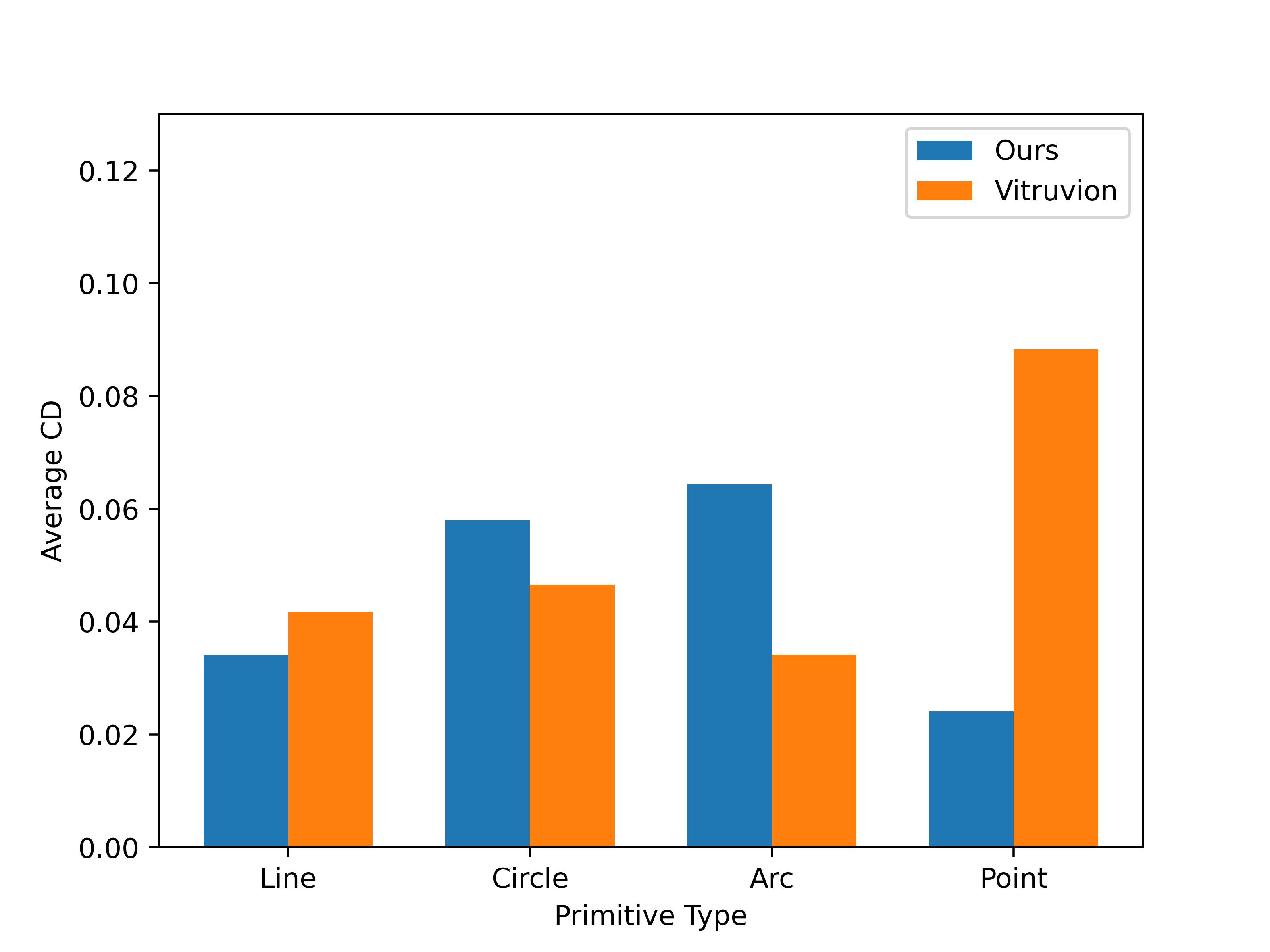}
    \caption{Comparison of the average CD for each primitives type.} 
    \label{fig:type dc}
\end{figure}

As seen in \figurename~\ref{fig:type dc}, compared with Vitruvion \cite{seff2021vitruvion}, the average CD between primitives of the same type. The better support of Vitruvion's \cite{seff2021vitruvion} auto-regressive model for the primitives' types represented in long sequential lists is reflected in the \figurename~\ref{fig:type dc}, where it can be observed that $Point$ has the largest CD, while $Arc$ has the smallest CD. Each primitive in PPI-Net is treated as an independent object, and there is a certain correlation between the number of parameters and the average CD of the primitives. Specifically, primitives with fewer parameters, such as $Point$, tend to have smaller average CD, while primitives with more parameters, such as $Arc$, typically have larger average CD. Additionally, the imbalance in the distribution of primitive types in the dataset affects the occurrence frequency of $Line$ and $Circle$, thereby influencing their average CD. As a result, $Line$ generally has a smaller CD compared to $Circle$.

\subsection{Import into CAD software}
\begin{figure*}[h!t]
    \centering
    \includegraphics[width=\textwidth,height=\textheight,keepaspectratio]{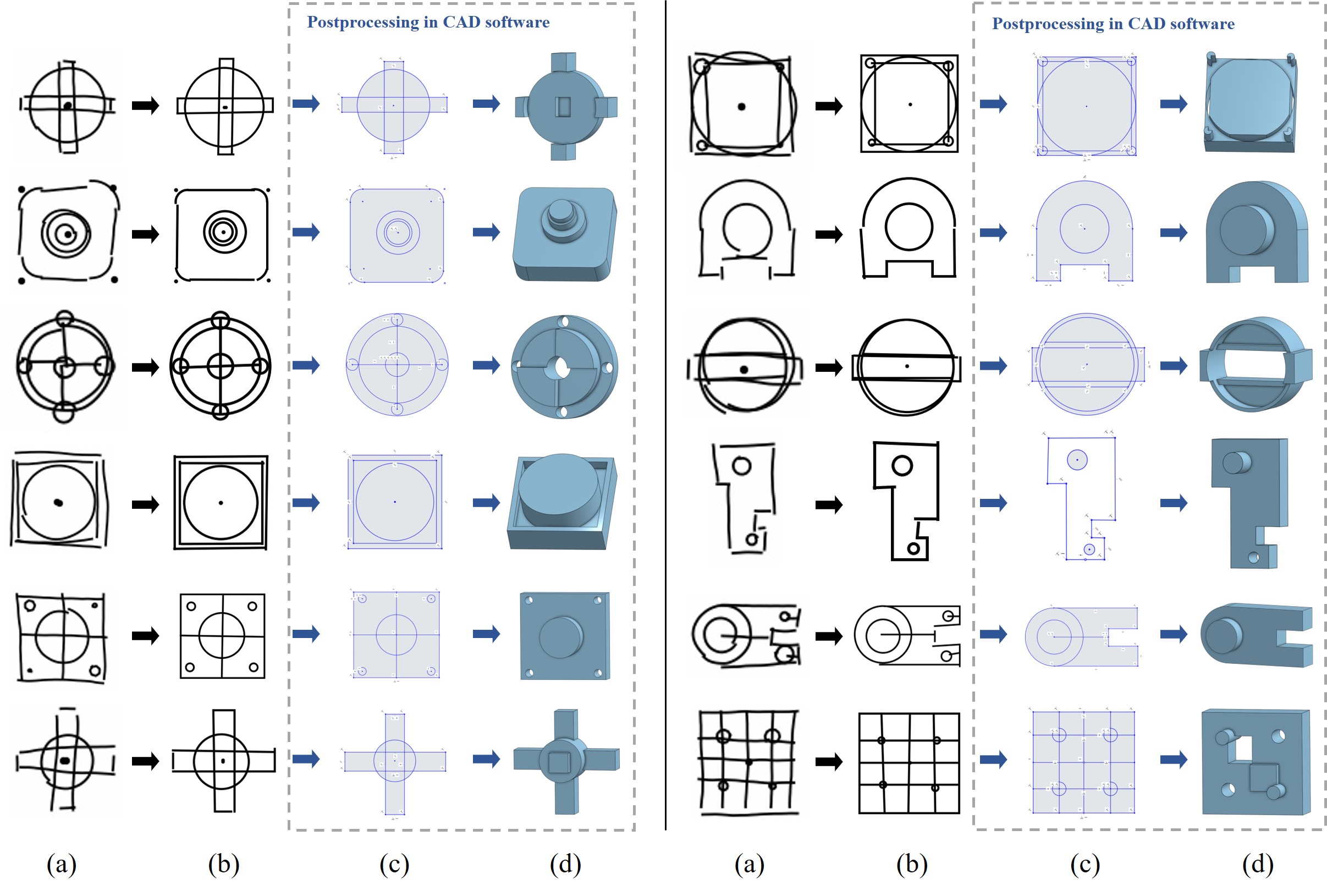}
    \caption{Editing infer primitives in CAD software. Hand-drawn sketch images $(a)$, infer parametric primitives $(b)$, import the parametric primitives, add constraints in Onshape Gui $(c)$, adjust primitives' positions, and use CAD operations to make the sketch to generate a 3D model $(d)$.} 
    \label{fig:contrast_2}
\end{figure*}

As shown in the \figurename~\ref{fig:contrast_2}, firstly, the hand-drawn sketch is input into the image primitive network. Subsequently, the extracted primitives are input into professional CAD software. Within the CAD software, various tools and commands can be utilized to add constraint relationships and adjust the positions and sizes of the primitives, ensuring the creation of a coherent sketch. These constraint relationships may involve coincident, parallel, perpendicular, and other conditions to meet design requirements and geometric specifications. Once the sketch achieves the desired coherence and geometric criteria, professional CAD operations like extrusion, hole-cutting, and rotation can result in the generation of the final CAD model.

\begin{table}[h!t]
\caption{Image primitive model evaluated on real hand-drawn sketches. The left side of the table represents our training strategy, while the top side represents the metrics. The downward arrow \textbf{$\downarrow$} indicates that smaller values are better, while the upward arrow \textbf{$\uparrow$} indicates that larger values are better. The metrics highlighted in bold indicate the optimal values.} 
\label{tab:table_two} 
\resizebox{\linewidth}{!}{ 
\begin{tabular}{l|cccc}
\hline
\multicolumn{1}{l|}{Training regimen} & \multicolumn{1}{c}{Type\_acc$\uparrow$} & \multicolumn{1}{c}{CD$\downarrow$} & \multicolumn{1}{c}{Precision$\uparrow$} & \multicolumn{1}{c}{Recall$\uparrow$}  \\ \hline
Precise rendering         & 82.5\%             & 0.0597      & 58.6\%              & 51.5\%           \\ 
Hand-drawn augmentation   & 92.6\%             & 0.0383      & 71.2\%              & 67.4\%           \\ 
Hand-drawn + affine       & \textbf{93.2\%}             & \textbf{0.0368}      & \textbf{73.9\%}              & \textbf{71.1\%}           \\ \hline     
\end{tabular}
}
\end{table}

\begin{table}[h!t]
\caption{Compare metrics with Vitruvion \cite{seff2021vitruvion}.} 
\label{tab:table_Four} 
\resizebox{\linewidth}{!}{ 
\begin{tabular}{c|cccccc}
\hline
\multicolumn{1}{c|}{Metric} & \multicolumn{1}{c}{Type\_acc$\uparrow$} & \multicolumn{1}{c}{CD$\downarrow$} & \multicolumn{1}{c}{Precision$\uparrow$} & \multicolumn{1}{c}{Recall$\uparrow$}  & \multicolumn{1}{c}{GFLOPs}& \multicolumn{1}{c}{Params} \\ \hline
Vitruvion \cite{seff2021vitruvion}                                     & 91.0\%             & 0.0428      & 66.9\%             & 65.3\%          & 230              & 172              \\ 
Ours                                    & \textbf{93.2\%}             & \textbf{0.0368}      & \textbf{73.9\%}             & \textbf{71.1\%}          & 93              & 61              \\ \hline
\end{tabular}
}
\end{table}

\begin{table}[h!t]
\caption{Ablation results for comparing the effects of denoising groups and the number of denoising groups in the same input conditions over 250 epochs, it is important to note that if the denoising groups are not used, the model not converge within the 250 epochs.} 
\label{tab:table_three} 
\resizebox{\linewidth}{!}{ 
\begin{tabular}{c|cccccc}
\hline
\multicolumn{1}{c|}{Denoise Groups} & \multicolumn{1}{c}{Type\_acc$\uparrow$} & \multicolumn{1}{c}{CD$\downarrow$} & \multicolumn{1}{c}{Precision$\uparrow$} & \multicolumn{1}{c}{Recall$\uparrow$} & \multicolumn{1}{c}{GFLOPs}& \multicolumn{1}{c}{Params} \\ \hline
No Group     & 80.2\%             & 0.0581       & 58.9\%             & 54.1\%          & 93              & 61              \\ 
1 Group   & 92.3\%             & 0.0388      & 67.9\%             & 66.1\%          & 93              & 61              \\ 
3 Groups     & \textbf{93.2\%}             & \textbf{0.0368}      & \textbf{73.9\%}             & \textbf{71.1\%}          & 93              & 61              \\ \hline           
\end{tabular}
}
\end{table}
\begin{table}[h!t]
\caption{Ablation results for denoising part. All models use 1 denoising group under the same default settings.} 
\label{tab:table_Five} 
\resizebox{\linewidth}{!}{ 
\begin{tabular}{ccc|cccc}
\hline
\multicolumn{1}{c}{\begin{tabular}[c]{@{}c@{}}Parameter \\ Denoise\end{tabular}} & \multicolumn{1}{c}{\begin{tabular}[c]{@{}c@{}}Label \\ Denoise\end{tabular}} & \multicolumn{1}{c|}{\begin{tabular}[c|]{@{}c@{}}Attention \\ Mask\end{tabular}} & \multicolumn{1}{c}{Type\_acc$\uparrow$} & \multicolumn{1}{c}{CD$\downarrow$} & \multicolumn{1}{c}{Precision$\uparrow$} & \multicolumn{1}{c}{Recall$\uparrow$} \\ \hline
$\checkmark$    & $\checkmark$     & $\checkmark$    & 92.3\%    & 0.0388      & 67.9\%       & 66.1\%                        \\
$\checkmark$    & $\ $     & $\checkmark$    & 91.9\%    & 0.0397      & 66.8\%       & 65.3\%                        \\                 
$\ $    & $\ $     & $\checkmark$    & 91.7\%    & 0.0403      & 66.5\%       & 65.1\%                        \\                 
$\checkmark$    & $\checkmark$     & $\ $    & 57.2\%    & 0.0730      & 31.2\%       & 26.7\%                        \\ \hline   
\end{tabular}
}
\end{table}

\subsection{Test strategy}

In Table \ref{tab:table_two}, we evaluate the performance of the image primitive model on a set of human-drawn sketches, which enable the above metrics evaluation section \ref{metrics}. These sketches were hand-drawn on a tablet computer in a 128×128-pixel bounding box.
We test three versions of the image primitive model on the hand drawings, each trained on a different type of rendering: precise renderings, renderings from the hand-drawn simulator, and renderings with random affine augmentations of the hand-drawn simulator. Both the hand-drawn simulation and augmentations substantially improve performance. 

\subsection{Ablation study}
We employ denoising groups to accelerate the convergence speed of the DETR \cite{DETR} model and improve the final metric scores. In Table \ref{tab:table_three}, we demonstrate the effects of denoising groups and varying the number of denoising groups on the results, under the same input conditions and with the same number of encoder and decoder layers. We aim to explore the denoising groups' effectiveness and determine the optimal configuration. We performed a set of ablation studies to assess the impact of each component and the results are presented in Table \ref{tab:table_Five}. The findings in Table \ref{tab:table_Five} demonstrate that each component utilized in the denoising training contributes to enhancing performance. It is worth highlighting that the absence of an attention mask, which prevents information leakage, leads to a significant decline in performance.

\section{Conclusion}
We propose an end-to-end network called PPI-Net for inferring parametric primitives from hand-drawn sketch images, which provides users with a more flexible and manipulable data representation. By employing parametric primitives inference, we can bind different parameters to various types of primitives, providing an effective means to optimize the representation of primitives and cater to real world usage scenarios. Ultimately, this approach helps engineers communicate and collaborate more efficiently with team members, clients, and suppliers.

\begin{figure}[t]
    \centering
\includegraphics[width=3.2in,height=4.8in,keepaspectratio]{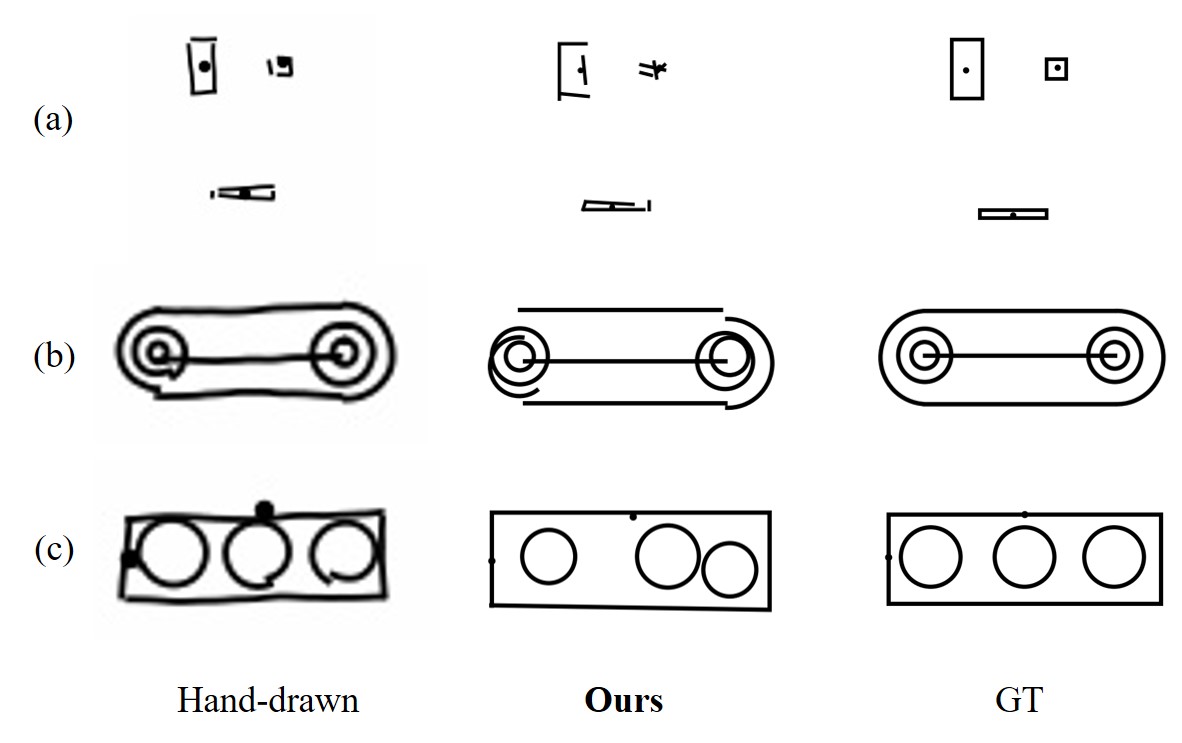}
    \caption{Failure cases about our image primitive model.} 
    \label{fig:failure}
\end{figure}

\textbf{ Limitation.} As shown in the (\figurename~\ref{fig:failure}($a$)), due to the fact that this method is derived from object detection, its inference capability for the small primitive is not strong. As shown in (\figurename~\ref{fig:failure}($b,c$)), there is a significant deviation between the extracted circles and arcs compared to GT. Therefore, we hope to further enhance the inference capability of image primitive model.

\textbf{ Future work.} Currently, the main method involves inferring primitives from hand-drawn sketch images, but it overlooks the constraints between primitives, which are pivotal for constructing continuous sketches. Therefore, in future work, we aim to achieve end-to-end inference primitives and the constraint relationships between them from hand-drawn sketch images in order to improve the quality and coherence of the generated sketches.

\section*{Acknowledgement}
We thank the anonymous reviewers for their valuable comments.
This work was supported in part by Natural Science Foundation of China (No. 62102328), 
in part by the Fundamental Research Funds for the Central Universities (No. SWU120076)
and  in part by the Open Project Program of State Key Laboratory of Virtual Reality Technology and Systems, Beihang University (No.VRLAB2023C01).

\bibliographystyle{plain}
\bibliography{main}

\end{document}